%% file: main.tex
\documentclass[letterpaper, 10 pt, conference]{ieeeconf}

\makeatletter
\let\NAT@parse\undefined
\makeatother
\IEEEoverridecommandlockouts                              
\usepackage{graphicx}
\usepackage{amsmath, amssymb}
\usepackage{booktabs, multirow, adjustbox}
\usepackage[table]{xcolor}
\usepackage[most]{tcolorbox}
\usepackage{algorithm, algorithmic}
\usepackage{url}
\usepackage[breaklinks,colorlinks,citecolor=blue]{hyperref}
\usepackage{caption} 
\usepackage{array}
\usepackage{makecell}
\usepackage{tabularx} 
\usepackage{balance}

\definecolor{highlightblue}{rgb}{0.8, 0.9, 1} 
\definecolor{ourrow}{rgb}{0.95, 0.95, 0.95}
\definecolor{speedcolor}{HTML}{77FA4C}
\definecolor{pathcolor}{HTML}{EA3323}

\title{\LARGE \bf
Latent-Centroid Steering: Single-Pass Classifier-Free Guidance for Command-Aligned Autonomous Driving

}

\newif\ifanonsub
\anonsubtrue       

\ifanonsub
  \author{Anonymous Authors}
\else
  \author{First Author$^{1}$, Second Author$^{2}$%
  \thanks{$^{1}$Affiliation 1. Email: xxx@xxx}%
  \thanks{$^{2}$Affiliation 2. Email: yyy@yyy}}
\fi

\author{Meibo Hu$^{1}$, Jiamian Wang$^{1}$, Pichao Wang$^{2}$, Zhiqiang Tao$^{1}$ \\
$^{1}$Rochester Institute of Technology, $^{2}$NVIDIA %
}

\begin{document}

\maketitle
\thispagestyle{empty}
\pagestyle{empty}

\begin{abstract}
Vision-language models (VLMs) have recently emerged as a promising paradigm for end-to-end autonomous driving, enabling agents to map multimodal inputs and high-level navigation instructions directly to executable trajectories. However, in practice, these models exhibit a persistent command-following gap: predicted trajectories often show weak sensitivity to navigation commands, resulting in incorrect behavior at critical decision points. We identify this issue as a form of conditional policy collapse, where regression-based training under multimodal trajectory distributions encourages the model to rely on dominant visual priors while marginalizing the language-conditioned signal. To address this issue, we introduce a principled formulation of classifier-free guidance (CFG) for regression-based vision-language driving. We show that CFG can be interpreted as isolating the instruction-induced residual in the action space by contrasting conditional and unconditional predictions, thereby explicitly amplifying the effect of the navigation command at inference time. However, a standard two-pass CFG introduces prohibitive latency for real-time control and produces noisy instance-level guidance directions.
Building on a mean-shift interpretation of CFG, we propose Latent-Centroid Steering (LCS), a single-pass guidance mechanism that replaces instance-level residuals with class-level latent shifts. By projecting conditional representations toward precomputed command-specific centroids, LCS performs class-level latent steering based on cluster geometry that is both more stable and computationally efficient. We demonstrate that LCS reduces inference latency by approximately 50\% while achieving stronger command adherence and improved driving performance on both closed-loop (Bench2Drive) and open-loop (nuScenes) benchmarks. Our results suggest that command following in VLA systems can be effectively improved through principled latent-space steering, providing a practical and theoretically grounded alternative to standard CFG for real-time autonomous driving. Code are
available at https://github.com/codingmlinprocess/LCS.
\end{abstract}

\section{INTRODUCTION}

The paradigm of autonomous driving is rapidly shifting toward end-to-end solutions, as demonstrated by recent breakthroughs in vision-language-action (VLA) models \cite{jiang2024senna, hwang2024emma,zhang2026minddriver}. Unlike traditional modular pipelines, a unified model maps raw visual observations and high-level language commands directly to low-level vehicle trajectories \cite{shao2024lmdrive,renz2025simlingo}. Unlike traditional modular pipelines, VLA-based planners leverage the vast internal knowledge and reasoning capacity of foundation models to achieve stronger semantic grounding and improved generalization to long-tail scenarios \cite{sima2024drivelm,tian2024drivevlm}.

Despite these advantages, a fundamental challenge persists: current VLA driving models can fail to reliably follow navigation commands \cite{renz2025simlingo,jiang2024senna}. For example, when instructed to “go straight,” the model may still follow a visually salient turning lane, or when instructed to “turn left,” it may continue along the dominant traffic flow. We refer to this phenomenon as the \emph{command-following gap}. While prior work attributes this to limitations of cross-modal fusion, we argue that the root cause also lies in the learning objective~\cite{hu2023planning,jiang2023vad,shao2024lmdrive}.

Specifically, trajectory prediction in end-to-end driving is typically formulated as a regression problem under multimodal target distributions \cite{hu2023planning,jiang2023vad,zheng2024genad}. Under such objectives, the optimal predictor tends to approximate the conditional mean trajectory given the visual context, which is dominated by high-frequency environmental priors such as lane geometry and traffic flow. As a result, the contribution of the language command becomes a low-magnitude perturbation that is easily ignored during training. This leads to a form of conditional policy collapse, where the learned policy effectively reduces to a visual-only policy even in the presence of explicit instructions \cite{renz2025simlingo,jiang2024senna,rawal2026nord,tang2026hermes}.

To address the above challenge, we introduce a principled formulation of classifier-free guidance (CFG)~\cite{ho2022classifier} for regression-based VLA driving. Originally developed for diffusion-based generative models, CFG effectively amplifies the difference between conditional and unconditional predictions. In the context of autonomous driving, we show that this difference (\emph{conditional vs unconditional}) isolates the instruction-induced residual in the action space, enabling the model to explicitly counteract dominant visual priors at inference time. By applying CFG to trajectory prediction, we transform the model from passively fusing modalities to actively steering its output according to navigation commands.

However, directly applying CFG introduces two major challenges. First, standard CFG requires two forward passes per control step—one for conditional inference and the other for unconditional — which doubles inference latency and is impractical for real-time driving. Second, the guidance direction derived from instance-level residuals is inherently noisy, as the unconditional prediction is itself a biased estimator of the marginal policy.

To overcome these limitations, we propose Latent-Centroid Steering (LCS), an efficient single-pass guidance mechanism derived from a mean-shift interpretation of CFG \cite{li2025towards}. Instead of computing instance-level residuals at inference time, LCS approximates the expected command-induced shift using precomputed class-conditional centroids within the latent feature space. This replaces noisy instance-level steering with stable cluster-level steering: the conditional latent representation is gently projected toward the semantic center of its command class, strengthening instruction alignment while preserving scene-specific perception.

Our study reveals that CFG and LCS represent two complementary forms of conditional steering. CFG performs instance-level steering by estimating per-sample residuals, while LCS performs class-level steering by leveraging the global geometric structure of the latent feature space \cite{li2025towards,sun2026latent}. In practice, we find that cluster-level steering not only eliminates the computational overhead of CFG but also yields more stable and accurate command-following behavior.

We validate our approach on both closed-loop and open-loop driving benchmarks. In particular, LCS achieves an $8\%$ relative gain in success rate over strong VLA baselines while maintaining real-time performance on Bench2Drive~\cite{jia2024bench2drive}; LCS also reduces trajectory prediction error by $5.6\%$ relative on nuScenes~\cite{caesar2020nuscenes}, demonstrating improved generalization to real-world data. Together, these results show that principled latent-space steering provides an effective solution to the command-following problem in VLA driving.
Our contributions are summarized as follows:
\begin{itemize}
    \item We formulate and analyze the command-following gap as a conditional policy collapse, providing new insights into regression-based VLA systems.
    \item We introduce CFG guidance for trajectory steering that explicitly isolates instruction-induced effects.
    \item We propose a novel Latent-Centroid Steering (LCS) inference approach, realizing a single-pass, cluster-level guidance mechanism that improves both efficiency and performance in real-time autonomous driving.
\end{itemize}

\section{RELATED WORK}
\subsection{Vision-Language Models for Autonomous Driving}
The evolution of autonomous driving has recently shifted from modular pipelines toward unified end-to-end neural architectures \cite{hu2023planning, jiang2023vad,mao2023language,sima2024drivelm,chen2024vadv2,yasarla2026generative,xiong2026unidrive,jia2025drivetransformer}. For open-loop planning, OpenDriveVLA \cite{zhou2025opendrivevla} emphasizes hierarchical alignment of 3D visual tokens, whereas Senna \cite{jiang2024senna} improves robustness by decoupling high-level intentions and low-level control. Complementing these specialized architectures, LLM-centric planners such as GPT-Driver \cite{mao2023gpt} and Emma \cite{hwang2024emma} demonstrate the potential of Large Language Models (LLMs) to act as zero-shot planners by reformulating driving as a token prediction task. Building on these foundations, LMDrive \cite{shao2024lmdrive} introduces a closed-loop system controlled by natural language instructions. SimLingo \cite{renz2025simlingo} extends this paradigm by incorporating a ``dreaming mode'' to strengthen language-action alignment within a real-time trajectory prediction framework. However, these methods typically rely on implicit cross-modal fusion mechanisms and do not provide explicit control over the influence of language commands, which may be  vulnerable to command-following failures in visually dominant scenarios \cite{jiang2024senna,renz2025simlingo,chen2026vilta}.

\subsection{Conditional Control and Guidance in Generative Models}Classifier-free guidance (CFG) has become a standard technique for improving condition adherence in generative models, particularly in diffusion-based image and video generation \cite{ho2022classifier,saharia2022photorealistic,ho2022video}. CFG works by contrasting conditional and unconditional predictions, effectively amplifying the influence of the conditioning signal. While CFG has been extensively studied in generative modeling, its application to regression-based decision-making systems such as autonomous driving remains largely unexplored. Moreover, the computational overhead of two-pass inference has limited its use in real-time control settings.

\subsection{Command Following and Policy Alignment}
The problem of aligning model outputs with high-level instructions has been studied in language models, robotics, and embodied AI. In driving, prior work~\cite{martinez2026natural} addresses this issue through architectural design, auxiliary losses, policy learning, or hierarchical planning structures \cite{shao2024lmdrive,jiang2024senna,zhou2025opendrivevla,li2025drive,zheng2025driveagent}. However, these approaches primarily focus on improving representation learning rather than explicitly controlling the conditional influence at inference time. Our work takes a complementary perspective by treating command following as a conditional policy alignment problem and introducing an explicit test-time steering mechanism.

\subsection{Latent Space Steering and Representation Geometry}
Recent studies in representation learning suggest that semantic attributes are often linearly organized in latent spaces, enabling manipulation via vector arithmetic or mean-shift operations \cite{li2025towards,wang2022modeling}. This idea has been explored in image generation, language models, and reinforcement learning policies. Our work builds on this line of research by introducing a centroid-based latent steering mechanism tailored to vision-language driving. Unlike instance-level residual guidance, our approach leverages class-level cluster structure to produce stable and efficient steering directions.

\begin{figure*}[t]
    \centering
    \includegraphics[width=1.0\linewidth]{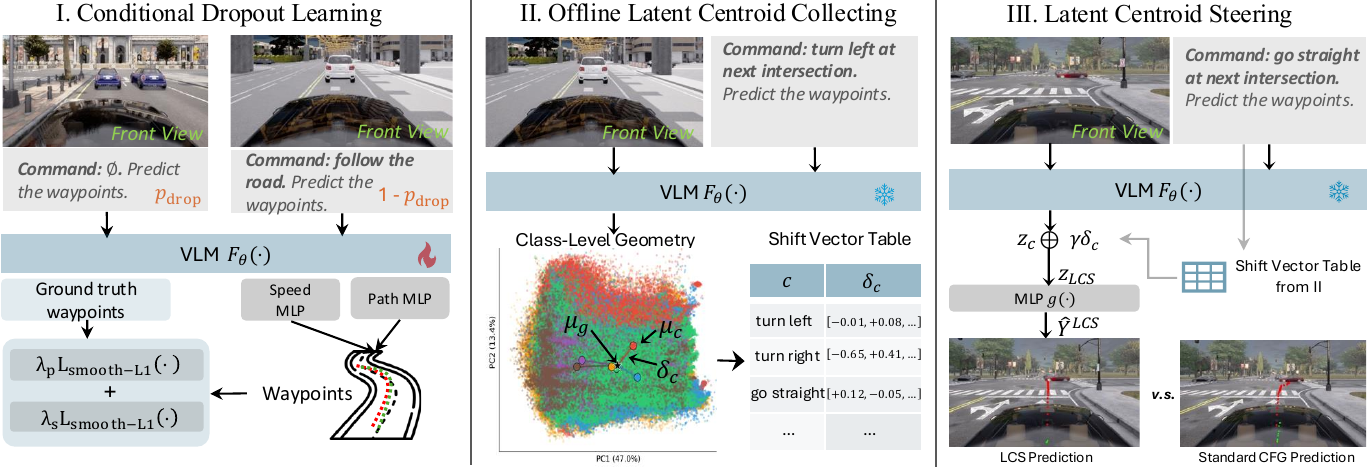}
    \caption{Overview of Latent-Centroid Steering (LCS). \textbf{Stage I}: The model is trained with conditional dropout, where the navigation command is randomly replaced with a null token with probability $p_{\text{drop}}$. \textbf{Stage II}: Command-specific centroids $\mu_c$ and the global centroid $\mu_g$ are computed offline from latent features, yielding a precomputed shift table $\{\delta_c\}$. \textbf{Stage III}: At inference, LCS applies the shift vector $\delta_c$ to the conditional latent $\mathbf{z}_c$ in a single forward pass, steering the prediction toward the intended command without requiring an additional unconditional pass.}
    \label{fig:framework}\vspace{-0.3cm}
\end{figure*}

\section{METHOD}
We present a framework for improving command following in VLA driving via latent-space steering. We first analyze the command-following gap as a form of conditional policy collapse in VLA systems. We then reformulate classifier-free guidance (CFG) as a residual steering mechanism for trajectory prediction. Finally, we derive Latent-Centroid Steering (LCS), a novel single-pass inference approach that enables stable and real-time command adherence.

\subsection{Problem Formulation}
At each timestep $t$, the driving model receives a visual observation $I_t \in \mathbb{R}^{H\times W \times 3}$ and a high-level language navigation command $c \in \mathcal{C}$ (e.g., \textit{turn left}, \textit{go straight}). The model predicts a sequence of future waypoints as
\begin{equation}
\hat{Y}_t = F_\theta(I_t, c), \quad \hat{Y}_t = \{y_{t+1}, \dots, y_{t+T}\},
\end{equation}
where $F_\theta(\cdot)$ denotes the VLA model. Training is performed with a regression loss against expert trajectories $Y_t^*$:
\begin{equation}
\mathcal{L} = \mathbb{E} \left[ \|F_\theta(I_t, c) - Y_t^*\|_1 \right].
\end{equation}
By minimizing this regression objective, the model converges toward the conditional mean of the expert trajectory distribution. Formally, the learned predictor approximates:
\begin{equation}
F_\theta(I,c) \approx \mathbb{E}[Y \mid I,c],
\end{equation}
where $I$ denotes the visual observation and $c$ the navigation command. To achieve a lower overall error, the model is encouraged to output the dominant mode in the training data. In driving datasets, the dominant mode corresponds to trajectories that are consistent with visual priors, such as lane structure and traffic flow. The model, therefore, learns to rely on these dominant patterns and becomes insensitive to the specific navigation command $c$, leading to:
\begin{equation}\label{eq: collapse}
F_\theta(I,c) \approx F_\theta(I,\emptyset),
\end{equation}
which we define as the \textbf{conditional policy collapse}, leading to the model producing similar trajectories regardless of commands. For example, a model may predict a lane-following trajectory even when the command explicitly requires a left turn at an intersection. In Eq.~\eqref{eq: collapse}, we denote $\emptyset$ as a null command, corresponding to the unconditioned prediction. 

\subsection{CFG as Residual Steering in Driving}
To counteract conditional collapse, we adopt classifier-free guidance (CFG)~\cite{ho2022classifier} and reinterpret it for trajectory prediction.
During training, we randomly drop the command with probability $p_{\text{drop}}$, so that the model jointly learns a conditional predictor $F_\theta(I,c)$ and an unconditional predictor $F_\theta(I,\emptyset)$. At inference, we obtain the guided trajectory by
\begin{equation}
\hat{Y}^{\text{cfg}} = F_\theta(I,\emptyset) +
w \left( F_\theta(I,c) - F_\theta(I,\emptyset) \right), 
\end{equation}
where the unconditional prediction $F_\theta(I,\emptyset)$ serves as a feasibility anchor that captures the physical constraints and statistical priors of the driving scene. We define the residual $\Delta(I,c) = F_\theta(I,c) - F_\theta(I,\emptyset)$ as the \textbf{instruction-induced action shift}. Since both predictions share the same network and visual input, the subtraction largely suppresses the shared visual components and isolates the contribution of the command $c$. The guidance scale $w \geq 1$ amplifies this command contribution to steer the trajectory toward the intended maneuver. This formulation is also supported by recent theoretical analysis~\cite{li2025towards}, which shows that CFG is equivalent to a mean-shift operation in representation space.

\textbf{Latent Space CFG.}
The above CFG formulation operates on the output waypoints. We can also apply the same principle in the model's intermediate feature space as shown in Fig.~\ref{fig:framework} \textit{left} part. Let $f_\theta$ denote the feature extractor that produces latent tokens before the prediction heads. The conditional and unconditional latent features are:
\begin{equation}
\mathbf{z}_c = f_\theta(I,c), \quad \mathbf{z}_0 = f_\theta(I,\emptyset).
\end{equation}
Latent-space CFG then performs the same linear extrapolation in this feature space:
\begin{equation}
\mathbf{z}_{\text{cfg}} = \mathbf{z}_0 + w(\mathbf{z}_c - \mathbf{z}_0).
\end{equation}
The steered features $\mathbf{z}_{\text{cfg}}$ are then decoded into waypoints through the prediction heads.

\textbf{Discussions on Standard CFG.}
While CFG provides a principled steering mechanism, two limitations hinder its direct application to real-time autonomous driving. First, standard CFG does not exploit the inherent structure of the VLA feature space. As we show in Sec.~\ref{sec:lcs}, navigation commands induce useful geometric properties that can be leveraged for more efficient steering. Second, CFG requires two full forward passes per timestep, which doubles the inference latency. This computational overhead is prohibitive for real-time control.

\begin{figure}[t]
    \centering
    \includegraphics[width=1.0\linewidth]{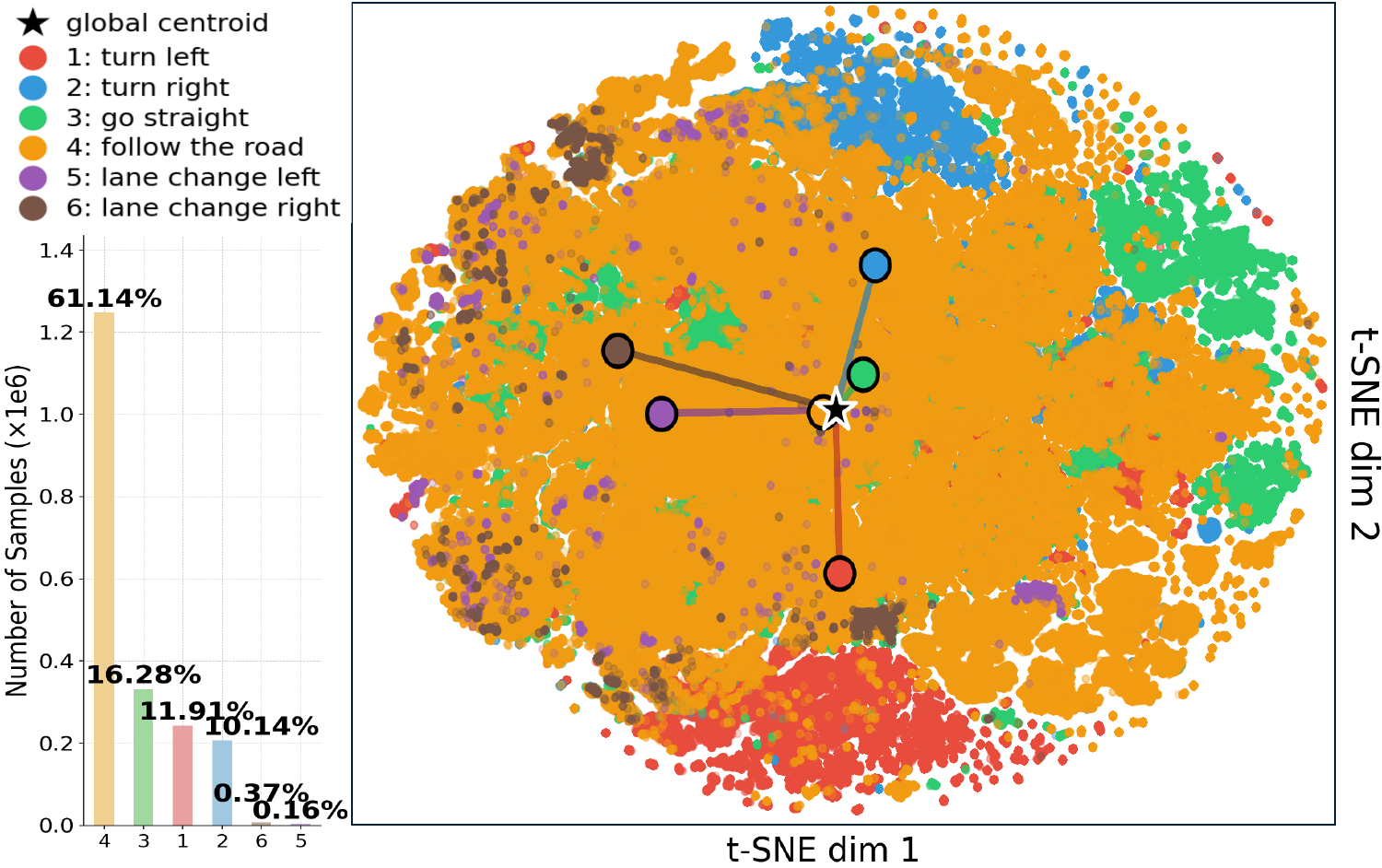}
    \caption{Visualization of command-conditioned latent features by t-SNE and the command distribution.} \vspace{-0.3cm}
    \label{fig:tsne}
\end{figure}

\subsection{Latent-Centroid Steering (LCS)}
\label{sec:lcs}
A key empirical finding of this work is that VLA models trained with command dropout exhibit a well-structured clustering property in the latent feature space: samples conditioned on the same navigation command form distinct groups in the feature space as shown in Fig.~\ref{fig:tsne}. Since the set of navigation commands is small and discrete, this clustering provides an opportunity to exploit class-level geometry for efficient steering.

\textbf{Centroid Definition.}
To capture the typical latent signature of each command, we define the command-specific centroid $\mu_c$ and the global centroid $\mu_g$ as
\begin{equation}
\mu_c = \mathbb{E}_{(I,c)}[f_\theta(I,c)], \quad
\mu_g = \mathbb{E}_{I,c}[f_\theta(I,c)].
\end{equation}
In practice, these expectations are estimated by averaging over training samples in a single offline pass:
\begin{equation}
\mu_c = \frac{1}{N_c} \sum_{i:c_i=c} f_\theta(I_i,c_i),
\quad
\mu_g = \frac{1}{N} \sum_i f_\theta(I_i,c_i).
\end{equation}
The centroid $\mu_c$ represents the prototypical latent state for command $c$, while $\mu_g$ represents the average latent state across all commands. As shown in the \textit{middle} of Fig.~\ref{fig:framework}, we define the shift vector as
\begin{equation}
\delta_c = \mu_c - \mu_g.
\end{equation}
This vector encodes the direction in latent space that distinguishes command $c$ from the overall data distribution. It captures how the feature representation should change to reflect the intent of a specific navigation command.

\textbf{LCS Formulation.}
During inference, LCS applies the precomputed shift vector $\delta_c$ to the conditional latent features:
\begin{equation}
\mathbf{z}_{\text{LCS}} = \mathbf{z}_c + \gamma \delta_c.
\end{equation}
The steered features are then decoded into the final trajectory:
\begin{equation}
\hat{Y}^{\text{LCS}} = g(\mathbf{z}_{\text{LCS}}),
\end{equation}
where $\mathbf{z}_c = f_\theta(I,c)$ is the conditional latent from a single forward pass, $g$ denotes the prediction heads, and $\gamma$ controls the steering strength as shown in Fig.~\ref{fig:framework} \textit{right} part.

\subsection{From CFG to LCS}
\textbf{LCS Interpretation.}
Three design principles underlie LCS. First, the shift vector $\delta_c$ represents a \textit{cluster-level direction}. Rather than computing a per-sample residual as in CFG, LCS uses the averaged geometric structure of the feature space to determine the steering direction. This yields a more stable signal. Second, LCS starts from the conditional feature $\mathbf{z}_c$ rather than the unconditional feature $\mathbf{z}_0$. The conditional feature already encodes the scene context attended through the lens of the given command. Starting from $\mathbf{z}_c$ preserves this scene understanding. Third, the shift acts as a \textit{gentle push} that moves $\mathbf{z}_c$ toward the centroid of its command cluster, strengthening the command signal without overriding the visual perception. We find that a proper $\gamma$ is sufficient to correct the command-following gap (see Sec.~IV). We provide the complete method in Algorithm~\ref{alg:lcs}.

\textbf{Relationship to CFG.}
Standard CFG steers the prediction using a per-sample residual from two forward passes:
\begin{equation}
\mathbf{z}_c - \mathbf{z}_0,
\end{equation}
where $\mathbf{z}_c$ and $\mathbf{z}_0$ need to be computed twice. By contrast, LCS replaces this with a precomputed class-level shift:
\begin{equation}
\underbrace{\mathbf{z}_c - \mathbf{z}_0}_{\text{CFG: per-sample residual}}
\quad \longrightarrow \quad
\underbrace{\mu_c - \mu_g}_{\text{LCS: centroid shift}}.
\end{equation}
This substitution brings three practical advantages. (1) LCS exploits the clustering property of the VLA feature space, using class-level centroids to provide a stable steering direction. (2) It replaces the per-sample residual estimation with an offline-computed shift vector, reducing variance across different driving scenes. (3) It requires only a single forward pass at inference, enabling real-time deployment.

\begin{algorithm}[t]
\caption{Latent-Centroid Steering (LCS)}
\label{alg:lcs}
\begin{algorithmic}[1]
\renewcommand{\algorithmicrequire}{\textbf{Input:}}
\renewcommand{\algorithmicensure}{\textbf{Output:}}
\REQUIRE Model $f_\theta$, prediction heads $g$, dataset $\mathcal{D}$, commands $\mathcal{C}$, dropout probability $p_{\text{drop}}$, steering strength $\gamma$
\ENSURE Guided trajectory $\hat{Y}^{\text{LCS}}$

\medskip

\STATE \textbf{// Stage I: Conditional Dropout Training}
\FOR{each training step}
    \STATE With probability $p_{\text{drop}}$, replace command $c$ with $\emptyset$
    \STATE Optimize $\mathcal{L} = \lambda_p \mathcal{L}_{\text{smooth-L1}}(\mathbf{p}, \hat{\mathbf{p}}) + \lambda_s \mathcal{L}_{\text{smooth-L1}}(\mathbf{s}, \hat{\mathbf{s}})$
\ENDFOR

\medskip

\STATE \textbf{// Stage II: Offline Centroid Collecting}
\FOR{each sample $(I_i, c_i) \in \mathcal{D}$}
    \STATE $\mathbf{z}_i \leftarrow f_\theta(I_i, c_i)$ \hfill $\triangleright$ \textit{Conditional forward pass}
\ENDFOR
\FOR{each command $c \in \mathcal{C}$}
    \STATE $\mu_c = \frac{1}{N_c}\sum_{i:c_i=c} \mathbf{z}_i$ \hfill $\triangleright$ \textit{Command centroid}
\ENDFOR
\STATE $\mu_g = \frac{1}{N}\sum_{i=1}^{N} \mathbf{z}_i$ \hfill $\triangleright$ \textit{Global centroid}
\FOR{each command $c \in \mathcal{C}$}
    \STATE $\delta_c = \mu_c - \mu_g$ \hfill $\triangleright$ \textit{Shift vector}
\ENDFOR

\medskip

\STATE \textbf{// Stage III: Latent-Centroid Steering (Online)}
\STATE $\mathbf{z}_c \leftarrow f_\theta(I, c)$ \hfill $\triangleright$ \textit{Single forward pass}
\STATE $\mathbf{z}_{\text{LCS}} \leftarrow \mathbf{z}_c + \gamma \cdot \delta_c$ \hfill $\triangleright$ \textit{Centroid steering}
\STATE $\hat{Y}^{\text{LCS}} \leftarrow g(\mathbf{z}_{\text{LCS}})$ \hfill $\triangleright$ \textit{Decode to waypoints}

\end{algorithmic}
\end{algorithm}

\section{EXPERIMENTS}

\input{main_table}

\subsection{Experimental Setup}
\textbf{Benchmarks and Datasets.} We evaluate our method on both closed-loop and open-loop datasets:
\begin{itemize}
    \item Closed-loop (Bench2Drive \cite{jia2024bench2drive}):
    The model is trained on the SimLingo-Data dataset \cite{renz2025simlingo} and deployed in the CARLA simulator \cite{dosovitskiy2017carla} across 220 routes (Town01--Town12). 
    To ensure a fair comparison of core navigation capabilities, we maintain the same base dataset as SimLingo \cite{renz2025simlingo} but restrict the training inputs to basic navigation commands and trajectory labels. Specifically, we do not utilize any supplementary data modalities from the original work, such as Q\&A pairs, driving commentaries, or future-state predictions (e.g., Dreamer mode).
    Performance is quantified via Driving Score (DS), Success Rate (SR), Efficiency, and Comfortness.
    \item Open-loop (nuScenes \cite{caesar2020nuscenes}): To evaluate trajectory prediction accuracy on real-world data, we report the Average Displacement Error (ADE).

\end{itemize}

\textbf{Navigation Commands.} We define a unified navigation command space $\mathcal{C}$ tailored to each benchmark. Specifically, for the SimLingo-Data \cite{renz2025simlingo} dataset, the space comprises six discrete high-level commands: \textit{Turn left}, \textit{Turn right}, \textit{Go straight}, \textit{Follow the road}, \textit{Lane change to the left}, and \textit{Lane change to the right}. For nuScenes \cite{caesar2020nuscenes}, we follow the standard protocol and utilize three primary categorical commands: \textit{Turn Left}, \textit{Turn Right}, and \textit{Go Straight}. For each dataset, the shift vectors $\boldsymbol{\delta}_c$ are precomputed strictly according to these specific command definitions to ensure precise alignment within the latent space.

\textbf{Implementation Details.} Our architecture employs an InternVL2-1B backbone with LoRA fine-tuning. The model is trained for 16 epochs with a batch size of 128. We adopt the AdamW optimizer with a cosine learning rate scheduler. Training is conducted on a server equipped with 8$\times$ NVIDIA A100 GPUs, utilizing Deepspeed for efficient distributed training. 
For the proposed LCS, the shift vectors $\boldsymbol{\delta}_c$ are precomputed from the training samples of SimLingo-Data (for Bench2Drive) and nuScenes (for nuScene evaluation), respectively, following prior works. This ensures that the latent steering remains grounded in the domain-specific feature distribution of each dataset. For closed-loop execution in Bench2Drive, we employ PID controllers to translate predicted waypoints $\mathbf{p}$ and speed $\mathbf{s}$ into vehicle control signals following prior works.

\subsection{Main Results on Bench2Drive}
Table~\ref{tab:main} reports the primary performance metrics for the closed-loop evaluation. We compare our proposed methods to various leading autonomous driving models. Overall, the proposed CFG and LCS guidance strategies achieve a clear performance gain over strong baselines.

\textbf{Quantitative Analysis.} As shown in Table~\ref{tab:main}, the prior work Simlingo achieves a Driving Score (DS) of $86.08\pm1.76$. By incorporating our proposed Latent-Centroid Steering (LCS), the DS increases to $87.18\pm0.52$, and the Success Rate (SR) improves significantly from $65.78\pm3.90$ to $71.16\pm1.23$. For all models, we report the mean and standard deviation across three independent training seeds.

\begin{table}[t]
\caption{Efficiency and performance comparison of guidance strategies. We evaluate models on the Bench2Drive closed-loop benchmark. Latency is measured by the averaged wall-clock time per decision cycle (from multimodal input to waypoint output) on a single NVIDIA H100 GPU.}
\label{tab:latency_comparison}
\begin{center}
\small
\begin{tabular}{lccc}
\toprule
Method & DS $\uparrow$ & SR $\uparrow$ & Latency $\downarrow$ \\
\midrule
SimLingo & 86.08$\pm$1.76  & 65.78$\pm$3.90  & 294.84ms \\
Action-Space CFG  & 86.87$\pm$1.58 & 70.57$\pm$2.43 & 522.44ms \\
Latent-Space CFG  & 87.07$\pm$0.86 & 70.91$\pm$2.20 & 563.55ms \\
\rowcolor{gray!15}
\textbf{LCS (Ours)} & \textbf{87.18$\pm$0.52} & \textbf{71.16$\pm$1.23} & \textbf{285.75ms} \\
\bottomrule
\end{tabular}
\end{center}

\end{table}

\textbf{Comparison with Two-pass CFG.} To ensure a fair comparison, we report the model-only latency, encompassing the entire forward pass from input embedding to predicted trajectory within one step in Table~\ref{tab:latency_comparison}. Compared to the two-pass CFG methods, which are the standard approach for conditional steering but necessitate double the inference budget, LCS significantly reduces inference latency by $\sim50\%$ with a single forward pass. While CFG provides a strong guidance signal by contrasting conditional and unconditional distributions, LCS achieves similar semantic steering by utilizing precomputed shift vectors in the latent space. This confirms that navigation commands can be effectively injected via a static latent shift, making LCS a highly efficient, low-latency option for the real-time autonomous driving.

\subsection{Main Results on nuScenes}
\textbf{Open-loop Evaluation on nuScenes.} To validate the real-world generalization of our method, we evaluate LCS on the nuScenes dataset. As shown in Table~\ref{tab:nuscenes_internal}, LCS reduces the average displacement error (L2) compared to the other methods. This improvement is achieved through a single forward pass, demonstrating that shifting the latent feature effectively corrects the model's ``statistical inertia" toward expert behaviors without redundant computations. The results suggest that LCS can successfully recover expert-like trajectories in real-world scenarios by steering the latent features toward the command-specific distribution.

\input{table_nuscenes_v2}

\begin{figure}[t]
    \centering
    \includegraphics[width=\columnwidth]{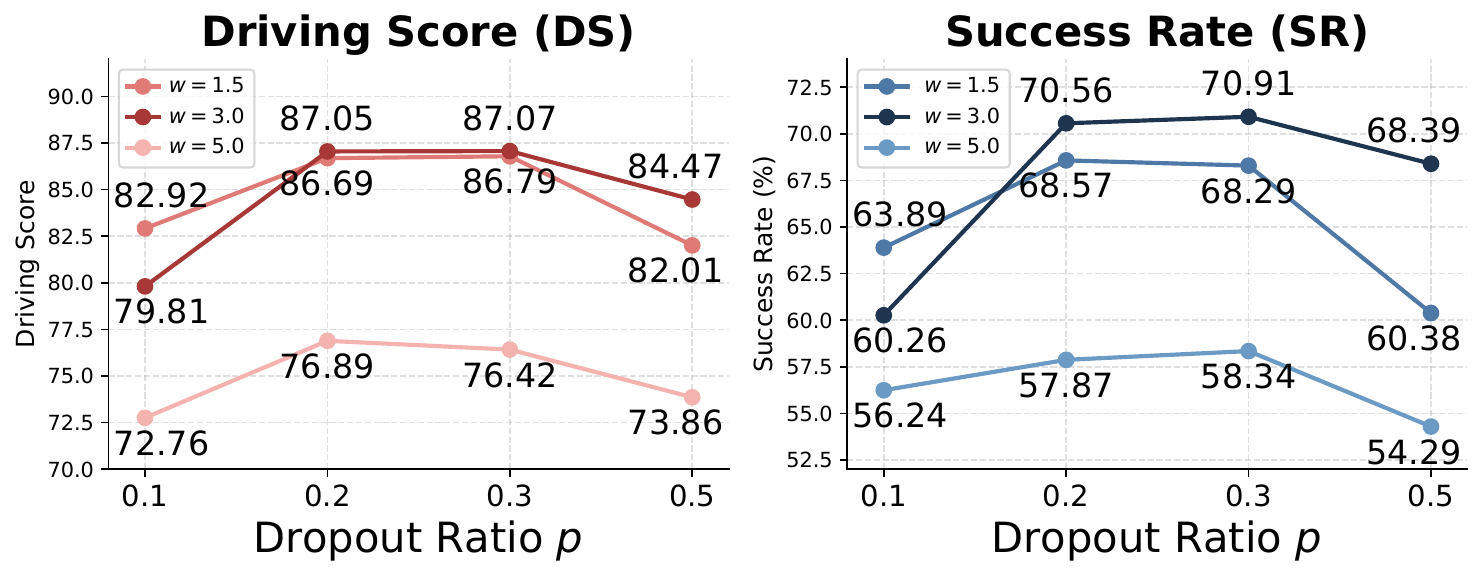}
    \caption{Dropout ratio and guidance scale analysis. We evaluate the performance with driving score   (left) and the success rate (right). Too samll/large $w$ or $p$ can cause lower results.  }
    \label{fig:dropout}
\end{figure}

\subsection{Model Discussion}
We conduct parameter analyses and ablation studies to discuss the effect of the proposed   designs the  key hyperparameter settings. All results are evaluated on the Bench2Drive  closed-loop simulations unless otherwise specified.

\textbf{Impact of Dropout Ratio $P_{\text{drop}}$.} 
The command dropout ratio during fine-tuning is critical for shaping the unconditional prior. As shown in Fig.~\ref{fig:dropout}, $P_{\text{drop}} = 0.3$ enables the best performance. A lower ratio results in an underdeveloped unconditional feature distribution, making the shift vector $\boldsymbol{\delta}_c$ less representative of the command intent. Conversely, an excessively high ratio ($P_{\text{drop}} > 0.5$) can undermine the model's command-following capability due to the insufficient conditional training.

\textbf{Guidance Scale $w$.} 
The guidance scale $w$ determines the overall magnitude of the shift applied to the latent features. As shown in Fig.~\ref{fig:dropout}, we find that $w = 3$ enables the best performance. When $w$ is too small, the navigation intent is not sufficiently emphasized, leading to degraded command-following ability. Conversely, excessively large values (e.g., $w > 5.0$) induce aggressive driving behaviors, such as abrupt steering and hard braking, as the model over-prioritizes the guidance signal at the expense of environmental perception.

\begin{table}[t]
\centering
\small
\caption{Ablation study on steering base across different scales $\gamma$ (with $p_{\text{drop}} = 0.3$). We evaluate LCS by steering from unconditional vs. conditional latent bases. }
\label{tab:base_gamma_comparison}
\setlength{\tabcolsep}{8pt} 
\begin{tabular}{l cc cc}
\toprule
 & \multicolumn{2}{c}{\textbf{$\gamma = 0.1$}} & \multicolumn{2}{c}{\textbf{$\gamma = 0.30$}} \\
\cmidrule(lr){2-3} \cmidrule(lr){4-5}
\textbf{Steering Setting} & \textbf{DS $\uparrow$} & \textbf{SR $\uparrow$} & \textbf{DS $\uparrow$} & \textbf{SR $\uparrow$} \\
\midrule
$\mathbf{z}_{uncond} + \gamma \Delta_{cmd}$ & 84.25 & 62.40 & 82.89 & 65.76 \\
\rowcolor{ourrow} \textbf{$\mathbf{z}_{cond} + \gamma \Delta_{cmd}$ (LCS)} & \textbf{87.18} & \textbf{71.16} & \textbf{86.42} & \textbf{68.15} \\
\bottomrule
\end{tabular}
\end{table}

\textbf{Ablation Study on Steering Base.} 
A key design choice in LCS is the selection of the base latent representation for steering. We investigate two configurations: (1) steering from the \textit{conditional} base, $\mathbf{z}_{cond} + \gamma \cdot \boldsymbol{\delta}_c$, and (2) steering from the \textit{unconditional} base, $\mathbf{z}_{uncond} + \gamma \cdot \boldsymbol{\delta}_c$. As shown in Table~\ref{tab:base_gamma_comparison}, using the conditional feature as the base yields superior stability. Since $\mathbf{z}_{cond}$ already resides within the proximity of the target command feature cluster, the shift vector $\boldsymbol{\delta}_c$ acts as a refinement that corrects the ``command-following gap". In contrast, steering from $\mathbf{z}_{uncond}$ requires a larger scale to reach the expert distribution, which introduces noise and leads to sub-optimal navigation performance.

\begin{figure}[t]
    \centering
    \includegraphics[width=\columnwidth]{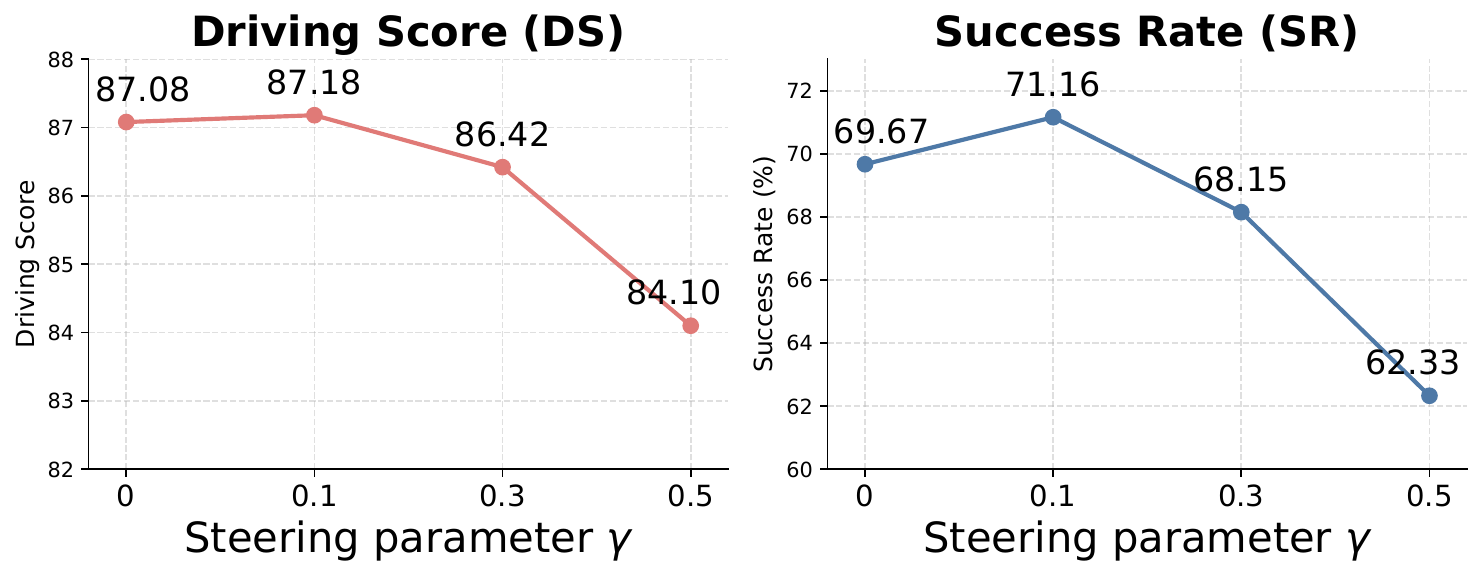}
    \caption{Discussion on the LCS steering parameter $\gamma$. Results are reported using the model trained with $p_{\text{drop}} = 0.3$. }
    \label{fig:gamma_abl}
\end{figure}

\begin{figure*}[t]
\centering
\begin{tabular}{@{}c@{\hspace{2pt}}c@{\hspace{5pt}}c@{\hspace{2pt}}c@{}}
Simlingo & LCS & Simlingo & LCS \\[1pt]
\includegraphics[width=0.24\textwidth]   {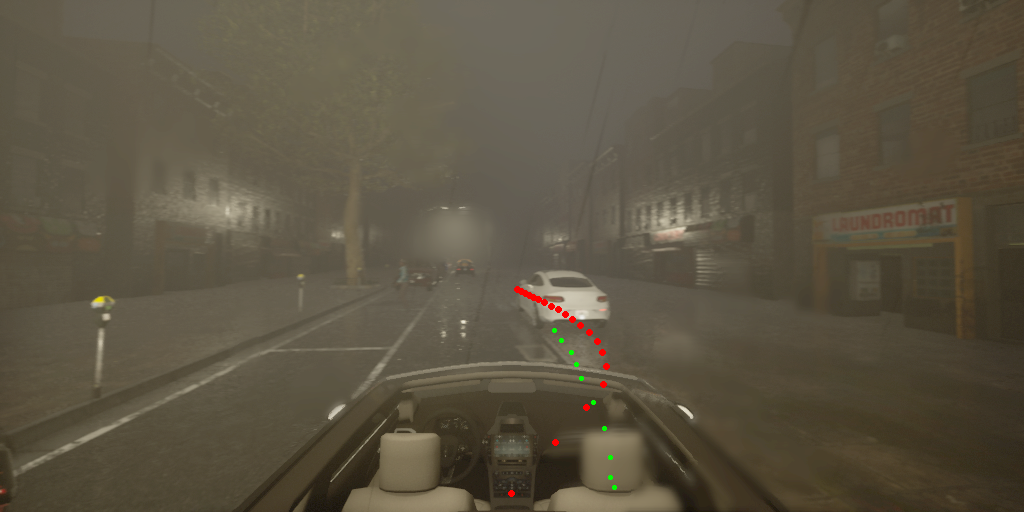} & \includegraphics[width=0.24\textwidth]{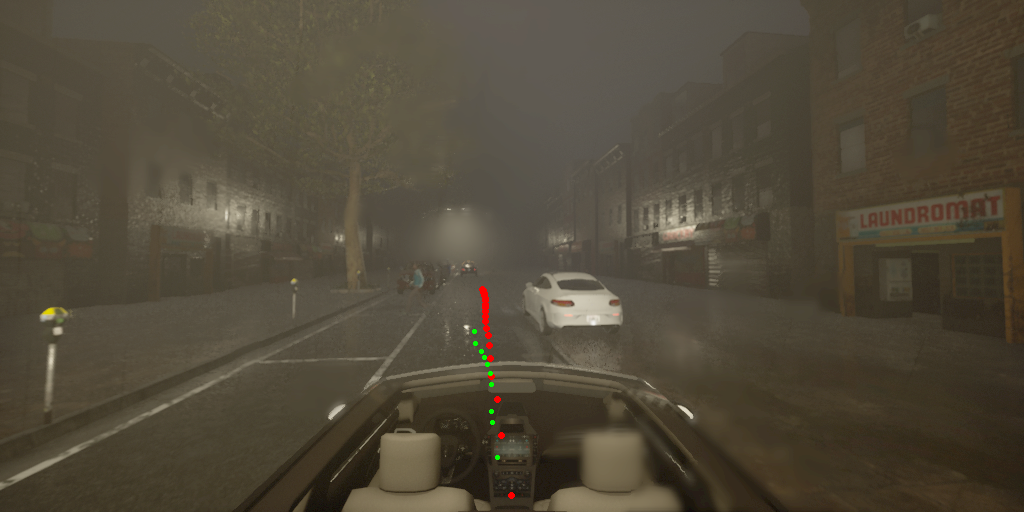} & \includegraphics[width=0.24\textwidth]{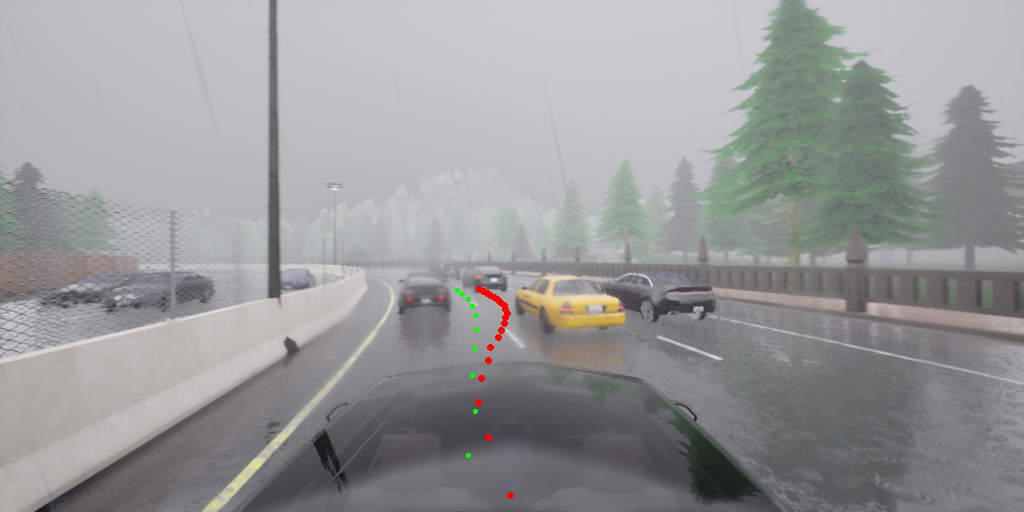} & \includegraphics[width=0.24\textwidth]{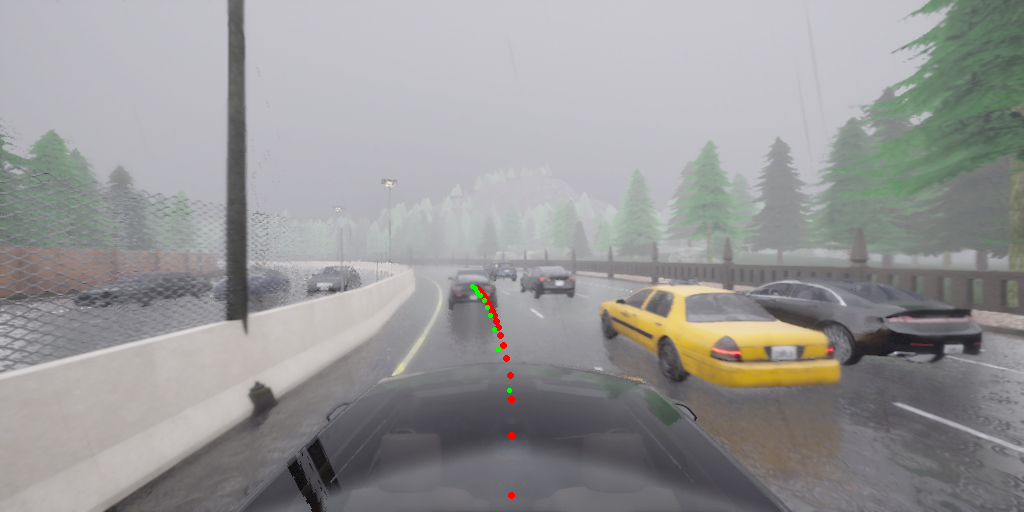}\\[1pt]

\multicolumn{4}{c}{``Follow the road."} \\[1pt]
\includegraphics[width=0.24\textwidth]{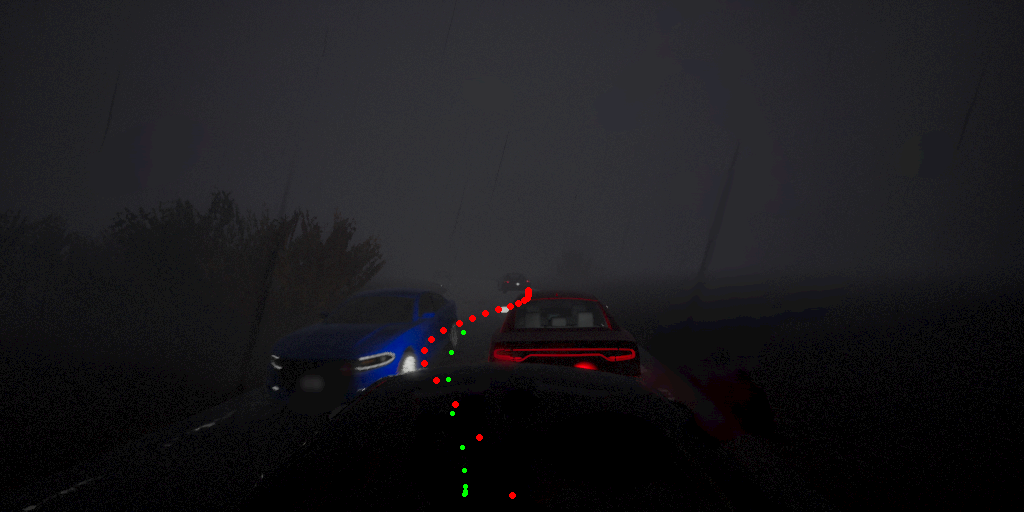} & \includegraphics[width=0.24\textwidth]{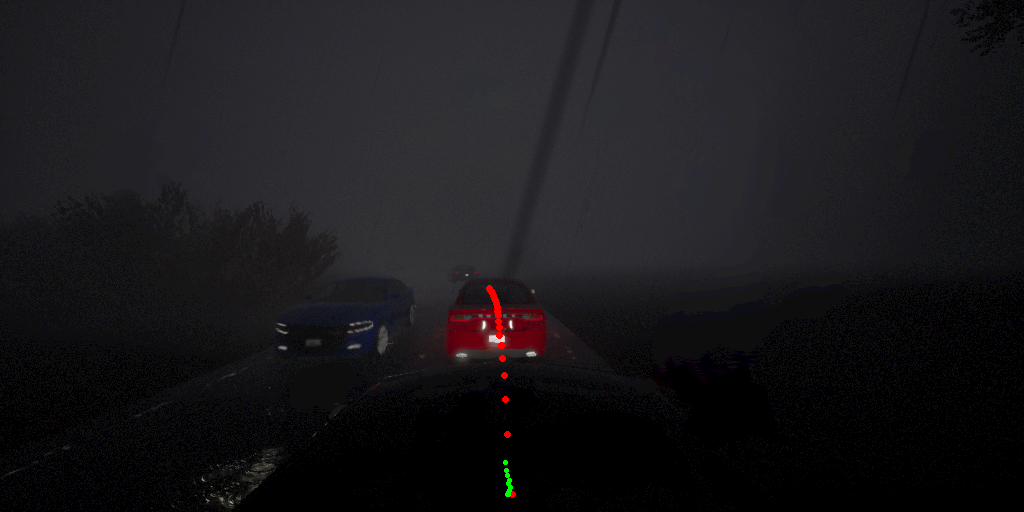} & \includegraphics[width=0.24\textwidth]{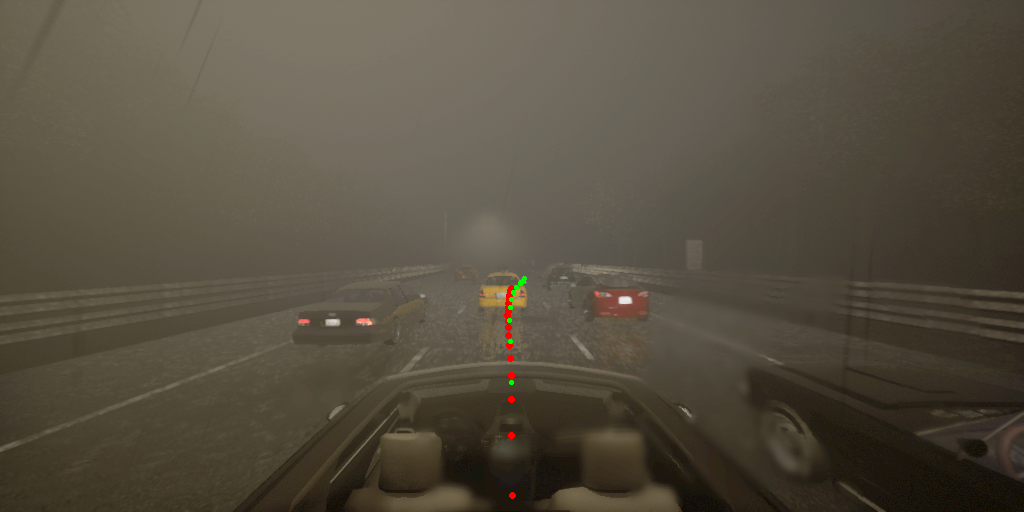} & \includegraphics[width=0.24\textwidth]{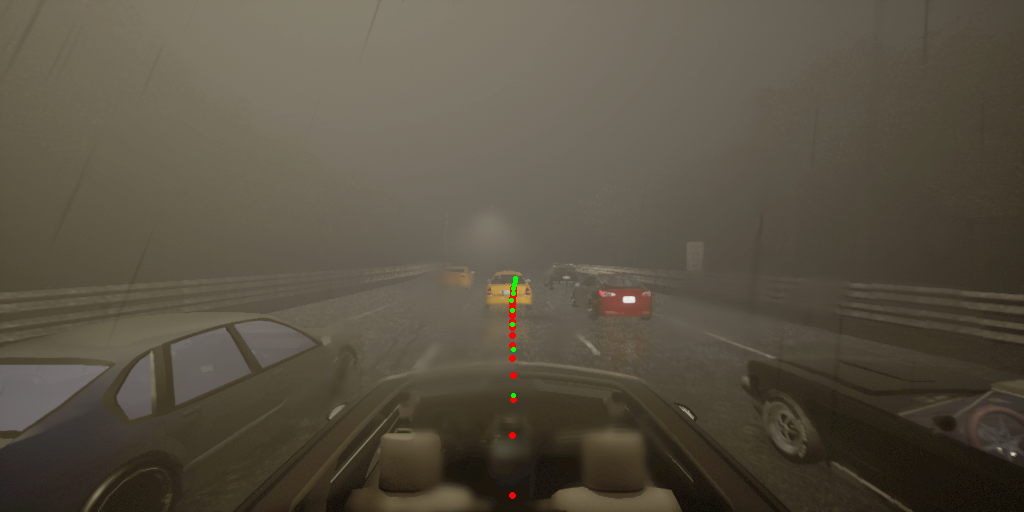}\\[1pt]
\multicolumn{4}{c}{``Follow the road."} \\[1pt]

\includegraphics[width=0.24\textwidth]{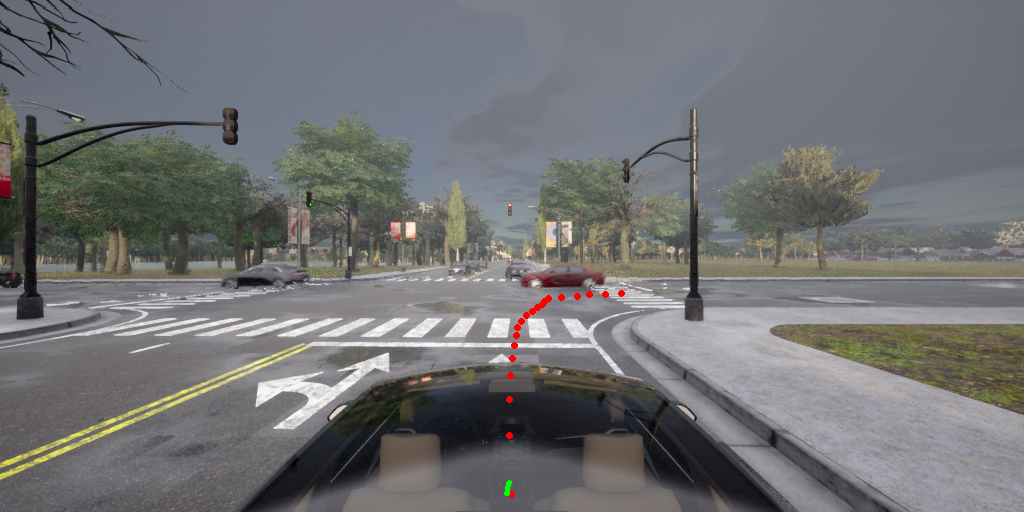} & \includegraphics[width=0.24\textwidth]{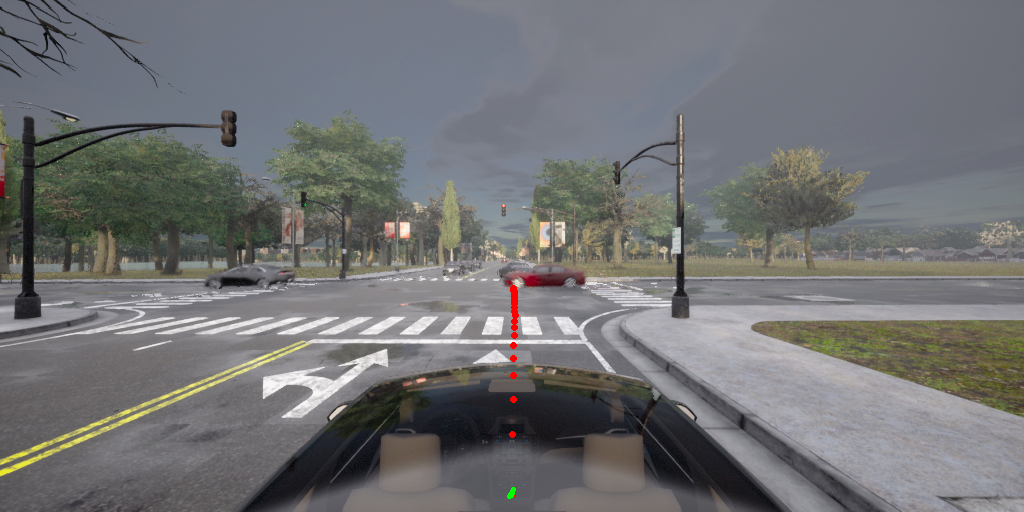} & \includegraphics[width=0.24\textwidth]{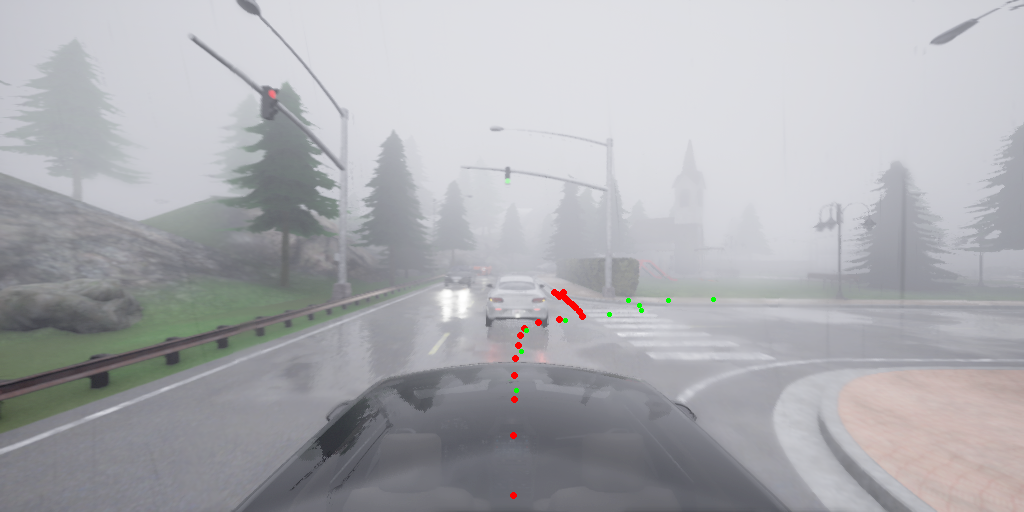} & \includegraphics[width=0.24\textwidth]{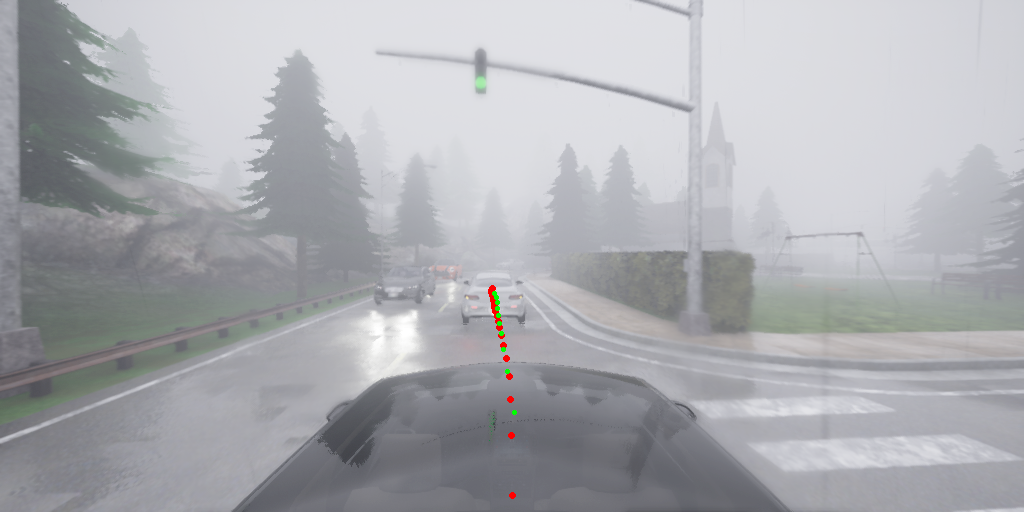}\\[1pt]
\multicolumn{4}{c}{``Go straight at the next intersection."} \\[1pt]
\includegraphics[width=0.24\textwidth]{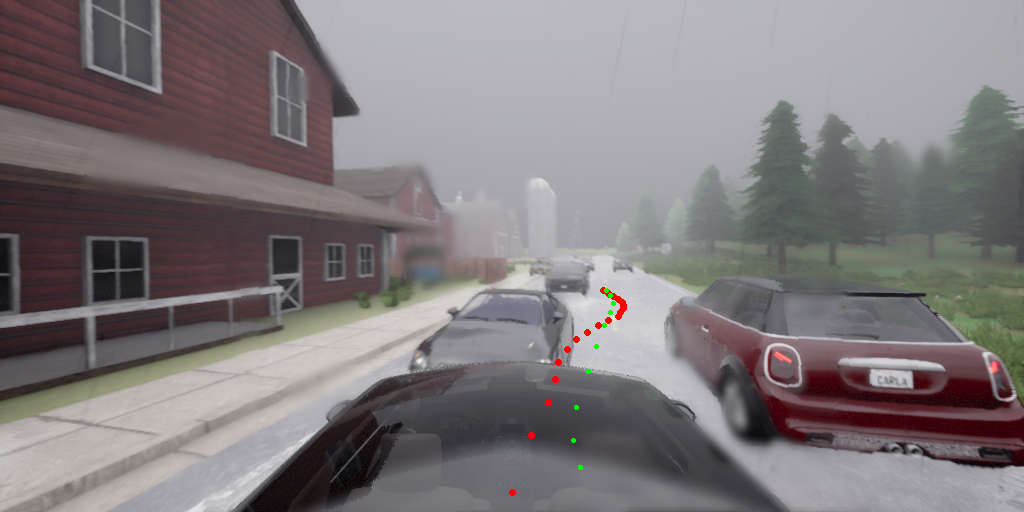} & \includegraphics[width=0.24\textwidth]{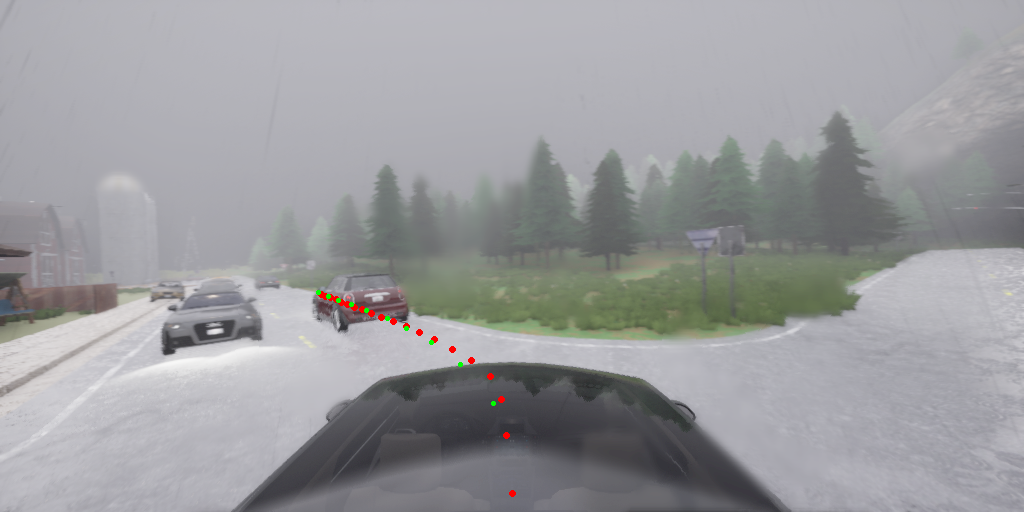} & \includegraphics[width=0.24\textwidth]{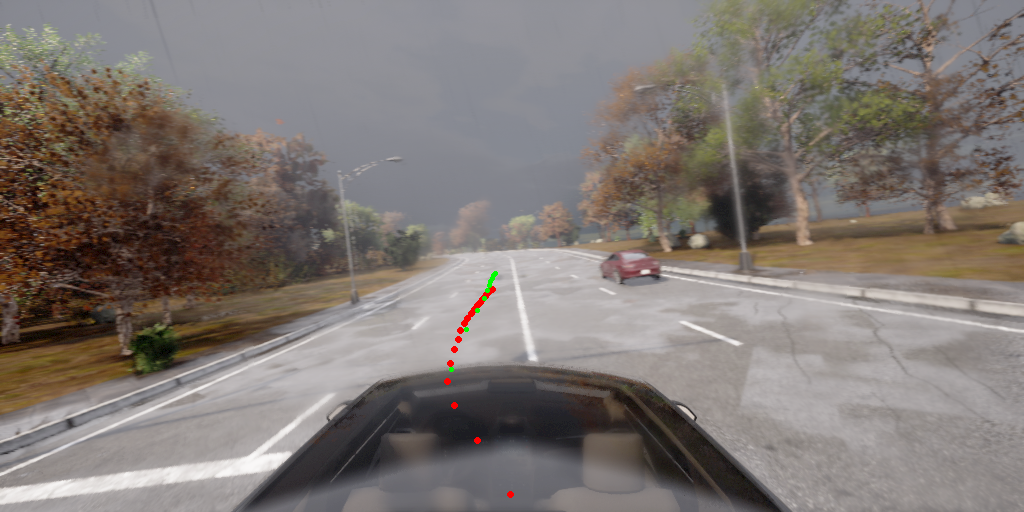} & \includegraphics[width=0.24\textwidth]{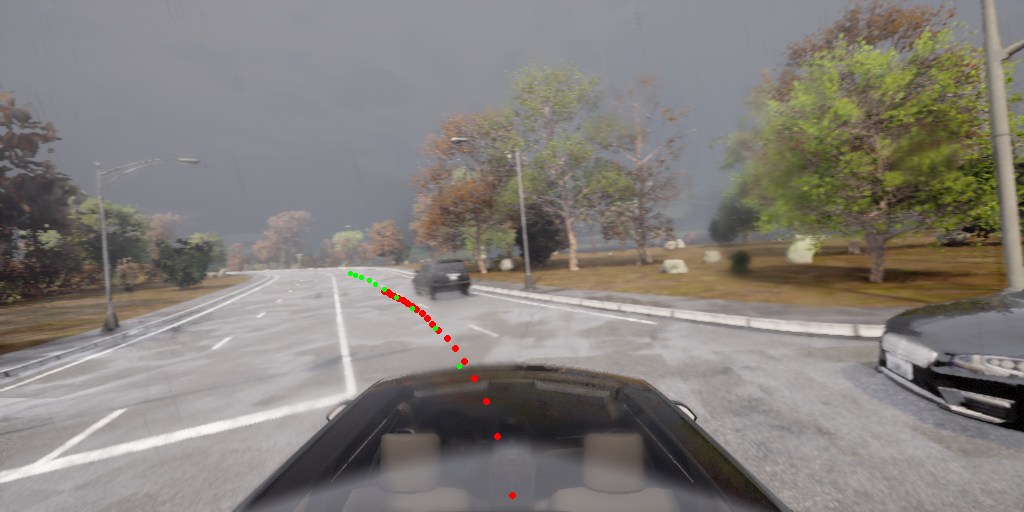}\\[1pt]
\multicolumn{4}{c}{``Turn left at the next intersection."} \\[1pt]
\includegraphics[width=0.24\textwidth]{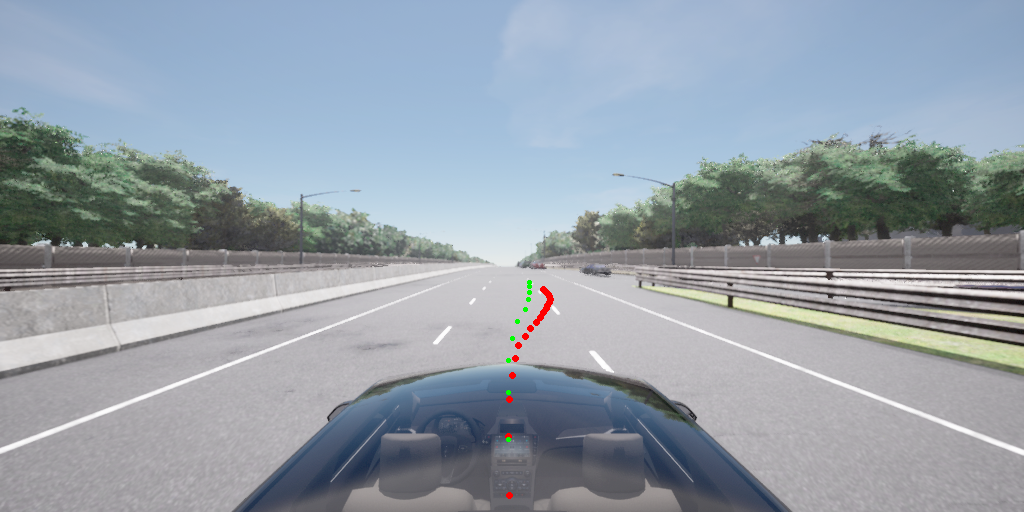} & \includegraphics[width=0.24\textwidth]{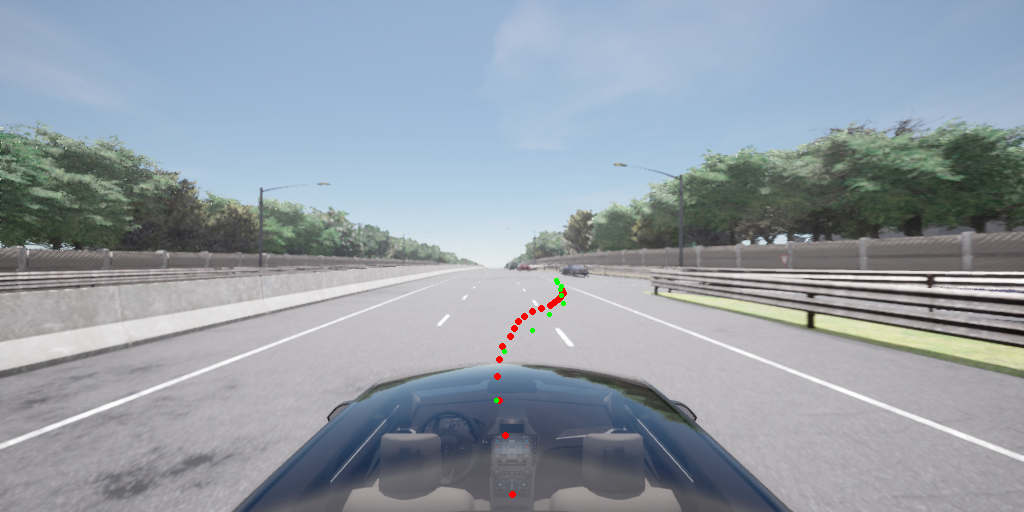} & \includegraphics[width=0.24\textwidth]{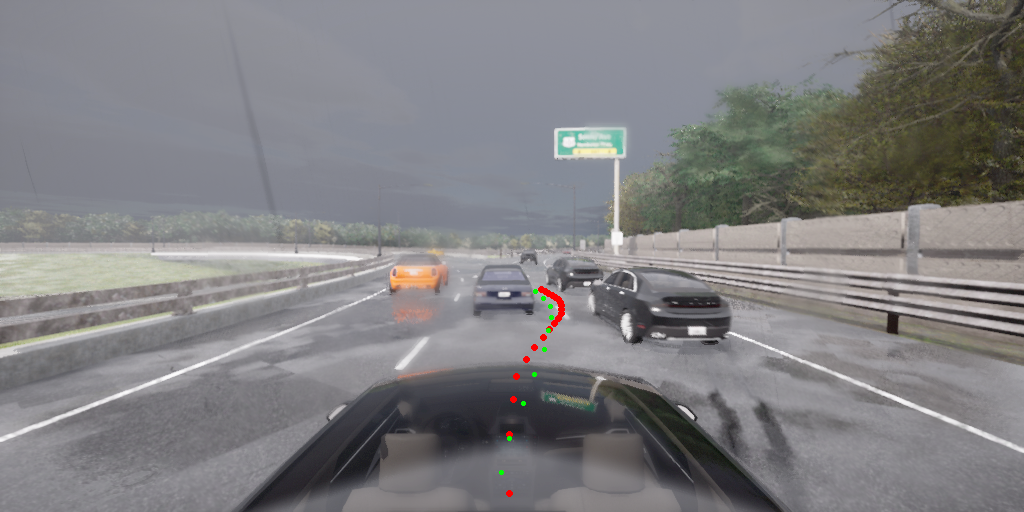} & \includegraphics[width=0.24\textwidth]{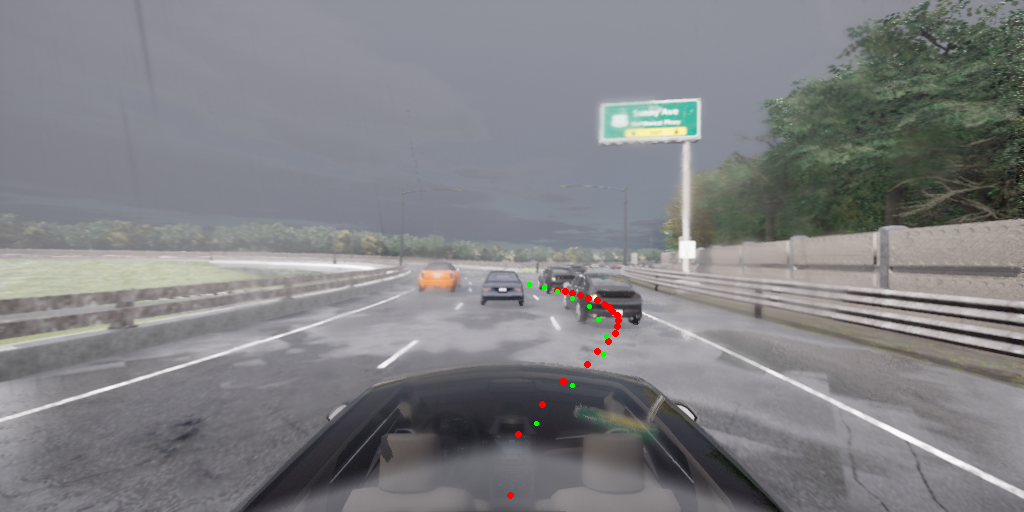}\\[1pt]
\multicolumn{4}{c}{``Do the lane change to the right."}

\end{tabular}
\caption{Qualitative comparison between LCS and Simlingo on Bench2Drive closed-loop evaluation by the CARLA simulator. Each row contains two prediction pairs guided by a single command.  In each image,  green dots denote predicted {\color{speedcolor}speed waypoints}, and red dots denote predicted {\color{pathcolor}path waypoints}. Both types of waypoints jointly guide the vehicle's driving behavior.}
\label{fig:full_comparison}\vspace{-0.2cm}
\end{figure*}

\textbf{LCS Steering Parameter $\gamma$.} 
The parameter $\gamma$ modulates the influence of the precomputed shift vector $\boldsymbol{\delta}_c$ when applied to the conditional base $\mathbf{z}_{cond}$. As shown in Fig.~\ref{fig:gamma_abl}, we find that a small value of $\gamma = 0.1$ achieves the superior performance. This suggests that while the vanilla conditional features follow the initial navigation intent, they can suffer from a subtle "statistical inertia" toward non-expert behaviors. A fine-grained shift ($\gamma=0.1$) is expected to steer the latent feature space toward the expert distribution.

\subsection{Qualitative Results on CARLA}
        Fig.~\ref{fig:full_comparison} presents a qualitative comparison between the proposed LCS and the Simlingo~\cite{renz2025simlingo}  on the Bench2Drive closed-loop simulator. As can be seen, LCS achieves better alignment between waypoint predictions and high-level language commands. For instance, under the command ``Follow the road" Simlingo misinterprets a temporarily stopped vehicle ahead as an obstacle to be bypassed, generating waypoints that deviate to an adjacent lane, whereas LCS correctly maintains the current lane and decelerates. For lane-change commands, LCS produces more precise waypoints that smoothly transition into the target lane. For ``Turn left/right'' commands at junctions, LCS predicts more precise turning points, whereas  Simlingo initiates the maneuver at incorrect locations, occasionally steering into opposing traffic or non-navigable areas. These results demonstrate that the proposed  method can effectively strengthen the alignment between language intent and waypoint predictions.

\section{CONCLUSION}
We studied the command-following gap in vision-language autonomous driving and identified it as a form of conditional policy collapse, where regression-based training causes policies to rely on dominant visual priors while underutilizing language instructions. To address this issue, we reformulated classifier-free guidance (CFG) as a residual steering mechanism for trajectory prediction and introduced Latent-Centroid Steering (LCS), a single-pass latent-space steering method that approximates CFG with class-level centroid shifts.
Experiments on both open-loop and closed-loop benchmarks showed that LCS outperforms vanilla VLAs and CFG-based steering in command adherence, and efficiency. These results highlighted the effectiveness of explicit test-time control of conditional influence for reliable vision-language driving.
Future work will extend LCS to complex, compositional instructions and adaptive steering mechanisms for broader autonomous driving systems.

\section{Acknowledgments}
This research was supported in part by the National Science Foundation under Grant No. 2502050 and the NVIDIA Academic Grant Program. The content is solely the responsibility of the authors and does not necessarily represent the official views of the funding agencies.

\balance
\bibliographystyle{IEEEtran}
\bibliography{main}

\end{document}

%% file: main_table.tex
\begin{table*}[t]
\centering
\caption{Main results on Bench2Drive~\cite{jia2024bench2drive} (Closed-loop evaluation). The proposed CFG and LCS guidance strategies consistently improve the camera-only (C) and multi-modal (C\&L) baselines on both the Driving Score (DS) and Success Rate (SR) metrics. Results are reported under Navigation Command (NC) and Target Point (TP) conditions.}

\label{tab:main}
\small
\begin{tabularx}{\textwidth}{l l l >{\centering\arraybackslash}X >{\centering\arraybackslash}X >{\centering\arraybackslash}X >{\centering\arraybackslash}X}
\toprule
\textbf{Method} & \textbf{Modality} & \textbf{Condition} & \textbf{DS $\uparrow$} & \textbf{SR $\uparrow$} & \textbf{Efficiency $\uparrow$} & \textbf{Comfortness $\uparrow$} \\ 
\midrule
    ThinkTwice \cite{jia2023think} & C & TP & 62.44 & 31.23 & 69.33 & 16.22 \\
    DriveAdapter \cite{jia2023driveadapter} & C\&L & TP & 64.22 & 33.08 & 70.22 & 16.01 \\ 
    SimLingo \cite{renz2025simlingo} & C & TP & 85.07$\pm$0.9 & 67.27$\pm$2.11 & 259.23$\pm$5.59 & 33.67$\pm$5.7 \\ \hline
    AD-MLP \cite{zhai2023rethinking} & C & NC & 18.05 & 0.00 & 48.45 & 22.63 \\
    UniAD-Tiny \cite{hu2023planning} & C & NC & 40.73 & 13.18 & 123.92 & 47.04 \\
    UniAD-Base \cite{hu2023planning} & C & NC & 45.81 & 16.36 & 129.21 & 43.58 \\
    VAD \cite{jiang2023vad} & C & NC & 42.35 & 15.00 & 157.94 & 46.01 \\
    GenAD \cite{zheng2024genad} & C & NC & 44.81 & 15.90 & -- & -- \\
    MomAD \cite{song2025don} & C & NC & 44.54 & 16.71 & 170.21 & 48.63 \\
    DriveTrans.-L \cite{jia2025drivetransformer} & C & NC & 63.46 & 35.01 & 100.64 & 20.78 \\ 
    ORION \cite{fu2025orion} & C & NC & 77.74 & 54.62 & 151.48 & 17.38 \\
    SimLingo \cite{renz2025simlingo} & C & NC & 86.08$\pm$1.76 & 65.78$\pm$3.90 & 259.23$\pm$5.59 & 33.67$\pm$5.72 \\ 
\midrule
\rowcolor{gray!15} 
\textbf{CFG (Ours)} & C & NC & \textbf{87.07$\pm$0.86} & \textbf{70.91$\pm$2.20} & \textbf{258.44$\pm$2.49} & \textbf{33.72$\pm$2.75} \\ 
\rowcolor{gray!15} 
\textbf{LCS (Ours)} & C & NC & \textbf{87.18$\pm$0.52} & \textbf{71.16$\pm$1.23} & \textbf{259.77$\pm$2.79} & \textbf{34.72$\pm$1.25} \\ 
\bottomrule
\end{tabularx}
\end{table*}

%% file: table_nuscenes_v2.tex
\begin{table}[t]
\centering
\small
\caption{Performance comparison on nuScenes. We adopt  Average L2 error and Collision rate to assess different methods.} 
\label{tab:nuscenes_internal}
\setlength{\tabcolsep}{8pt}
\begin{tabular}{l c c}
\toprule
\textbf{Method} & \textbf{Avg. L2 (m) $\downarrow$} & \textbf{Collision (\%) $\downarrow$} \\
\midrule
UniAD \cite{hu2023planning} & 1.03 & 0.31 \\
OccWorld-T \cite{zheng2024occworld} & 1.52 & 0.70 \\
ST-P3 \cite{hu2022st} & 2.11 & 0.71 \\
GenAD \cite{zheng2024genad}  & 0.91 & 0.43 \\
\midrule
Simlingo & 0.90 & 0.44 \\
\rowcolor{ourrow} \textbf{LCS (Ours)} & \textbf{0.85} & \textbf{0.41} \\
\bottomrule
\end{tabular}
\end{table}